\documentclass[12pt,twoside]{article}
\usepackage[paperwidth=6.75in,paperheight=10in,inner=20mm,outer=30mm,top=30mm,bottom=40mm]{geometry}
\usepackage{fontspec,longtable,booktabs,array,calc,graphicx,multirow,pdflscape,ragged2e,enumitem,tocloft,titlesec,xurl}
\usepackage{hyperref}
\usepackage{bookmark,orcidlink}
\usepackage{footnotehyper}
\makesavenoteenv{longtable}
\newfontfamily\footnotelabelfont[Path=fonts/,UprightFont=PlayfairDisplay-Bold.ttf,Numbers=OldStyle]{Playfair Display}
\newfontfamily\endmatterfont[Path=fonts/,UprightFont=EBGaramond-Italic.otf]{EB Garamond Italic}
\usepackage{fancyhdr,etoolbox}
\fancypagestyle{plain}{\fancyhf{}\fancyfoot[LE,RO]{\fontsize{11pt}{11pt}\selectfont\thepage}}
\makeatletter
\renewcommand\normalsize{\@setfontsize\normalsize{11.5pt}{14pt}}\normalsize
\renewcommand\footnotesize{\@setfontsize\footnotesize{10pt}{12pt}}
\ExplSyntaxOn
\cs_new_protected:Npn \ylj_digits:n #1 {
 \tl_map_inline:nn {#1} {
  \str_case:nnF {##1} {
   {3}{\raisebox{-1pt}[0pt][0pt]{3}} {4}{\raisebox{-1pt}[0pt][0pt]{4}}
   {5}{\raisebox{-1pt}[0pt][0pt]{5}} {7}{\raisebox{-1pt}[0pt][0pt]{7}}
   {9}{\raisebox{-1pt}[0pt][0pt]{9}}
   {6}{\raisebox{1pt}[0pt][0pt]{6}} {8}{\raisebox{1pt}[0pt][0pt]{8}}
  } {##1}
 }
}
\cs_new_protected:Npn \yljdisplaydigits {\exp_args:Ne \ylj_digits:n {\@thefnmark}}
\ExplSyntaxOff
\newcommand\yljfootnotelabel{\yljdisplaydigits.}
\renewcommand\@makefnmark{\hbox{\@textsuperscript{\normalfont\@thefnmark}}}
\renewcommand\@makefntext[1]{\parindent0pt\noindent\RaggedRight\makebox[1.9em][l]{{\footnotelabelfont\yljfootnotelabel}}\hangindent=1.9em\hangafter=1 #1}
\makeatother
\renewcommand\footnoterule{\kern0pt\hrule width\columnwidth height .4pt\kern-.4pt}
\setlist{nosep,leftmargin=1.2em}
\titleformat{\section}{\sffamily\bfseries\fontsize{10pt}{12pt}\selectfont\addfontfeatures{LetterSpace=13}\raggedright}{}{0pt}{}
\titleformat{\subsection}{\sffamily\bfseries\fontsize{9pt}{11pt}\selectfont\addfontfeatures{LetterSpace=13}\raggedright}{}{0pt}{}
\titleformat{\subsubsection}{\sffamily\bfseries\fontsize{9pt}{11pt}\selectfont\addfontfeatures{LetterSpace=13}\raggedright}{}{0pt}{}
\titlespacing*{\section}{0pt}{1.4em}{.55em}
\titlespacing*{\subsection}{0pt}{1em}{.35em}
\AtBeginEnvironment{longtable}{\fontsize{7pt}{8.4pt}\selectfont\setlength\tabcolsep{2pt}\RaggedRight}
\hypersetup{hidelinks,pdfcreator={XeLaTeX via Pandoc AST}}

\usepackage{float}
\usepackage{newunicodechar}
\newfontfamily\glyphfallback[Path=fonts/,UprightFont=DejaVuSans.ttf]{DejaVu Sans}
\newunicodechar{⇔}{{\glyphfallback ⇔}}

\hypersetup{pdftitle={From Deliberately Constructed Systemic Representations of the World of
Activity to Dialogue: A Five-Level Communication Architecture for Strong
Language-Based ASI},pdfauthor={Maxim Osovsky},pdfkeywords={Ai, AGI, ASI, SI, Artificial Intelligence, Artificial Superintelligence;
Artificial General Intelligence, Super Intelligence, activity-based
communication; Meaning--Text Theory; generative grammar; discourse
comprehension; systems approach; systemic representations.},pdfsubject={}}
\begin{document}\normalsize\sloppy
{\hyphenpenalty=10000\exhyphenpenalty=10000\raggedright\fontsize{16pt}{19.75pt}\selectfont\noindent From Deliberately Constructed Systemic Representations of the World of
Activity to Dialogue: A Five-Level Communication Architecture for Strong
Language-Based ASI\par}\vspace{16pt}

\noindent\begin{minipage}[t]{\linewidth}\centering
{\fontsize{12pt}{12pt}\selectfont\mbox{Maxim Osovsky\,\raisebox{1mm}{\orcidlink{0009-0004-1594-0940}}}\par}
{\fontsize{10pt}{12pt}\selectfont Adam University\par}
{\fontsize{12pt}{12pt}\selectfont Bishkek\par}
\end{minipage}\par\vspace{16pt}

\begingroup\fontsize{10pt}{13.5pt}\selectfont\noindent{\sffamily\bfseries\fontsize{7.7pt}{9.1pt}\selectfont\addfontfeatures{LetterSpace=13}ABSTRACT.}\ In working communication with models and agents of strong language-based
ASI, users must be able to understand the basic representations of the
world, activity, the economy, and economic organization on which those
models and agents rely. The central thesis of this paper is that, at
present, a language model's context is fundamentally incapable of
presenting the foundations of utterance formation for examination and
verification: these foundations are established in advance by acquired
schemas, a systemic representation of the world of activity, and the
connections accepted as true within it. Neither context nor trained LLM
weights, in themselves, attest to the presence of deliberately
constructed systemic representations of the world of activity. The paper
proposes five levels of content organization for strong language-based
ASI: deliberately constructed systemic representations of the world of
activity, a situation as a fragment of those representations, its
narrative description, the full sense of a message, and text. A
situation's schema-model is not identical to a story about it. A
recipient can reconstruct the full sense of a message from an
abbreviated text through jointly constructed and agreed-upon
representations and the situation of joint activity. Reverse work is
integral to the architecture: proceeding from the full sense of the
message through an extended description of the situation and its
schema-model to the reconstruction of systemic foundations. The
theoretical foundations draw on G. P. Shchedrovitskiy's systems
methodology, A. Korzybski's structural differential, and I. A. Mel'čuk's
Meaning--Text theory. The author's contribution is a specific
systemic-activity organization of content that makes it possible to
examine and adjust the foundations of a message and to control the
consequences of changes to them. Further research concerns the
construction of large graphical models of activity on reconstructible
systemic foundations, as distinct from the generation of text from
context.\par\endgroup

\par\vspace{6pt}\begingroup\rmfamily\mdseries\fontsize{10pt}{13.5pt}\selectfont\noindent\textit{Keywords:} Ai, AGI, ASI, SI, Artificial Intelligence, Artificial Superintelligence;
Artificial General Intelligence, Super Intelligence, activity-based
communication; Meaning--Text Theory; generative grammar; discourse
comprehension; systems approach; systemic representations.\par\endgroup

\clearpage\begingroup\fontsize{11.5pt}{14pt}\selectfont\tableofcontents\endgroup\clearpage

\section*{1. Position: What Is the Foundation of a Language Model's Response?}
\phantomsection
\label{yale-heading-002}
\addcontentsline{toc}{section}{1. Position: What Is the Foundation of a Language Model's Response?}

In human communication, a response is grounded in acquired schemas, a
systemic representation, and the connections accepted as true within it,
rather than in the mere presence of some context. P. G. Shchedrovitskiy
formulates the distinction between presenting a representation and
mastering it: ``The world must not only be presented in the forms of
ideal representations and ontological pictures; it must also be brought
within the sphere of human practical activity in its real
forms''.\footnote{P. G. Shchedrovitskiy, \emph{Философия развития и
  конструктивная психология {[}Philosophy of Development and
  Constructive Psychology{]}}, in Педагогика самоопределения и
  гуманитарные практики {[}The Pedagogy of Self-Determination and
  Humanistic Practices{]} 38 (Barnaul, 1999) (pp.~38--48; item 9),
  \url{https://shchedrovitskiy.com/phylosophiya-razvitiya-i-konstruktivnaya-psychologiya}.
  \emph{Unless otherwise indicated, translations of quotations from
  Russian sources are by the author.}} In this paper, a foundation is an
organization of representations that establishes distinctions between
processes and structures, specifies their connections, and determines
the permissible content of a message in joint activity.

What comes first in forming a description of activity: the context from
which structural and systemic representations emerge, or already
acquired, albeit incomplete, representations that determine how that
context is constructed? The proposed formulation requires consideration
of both directions: elaborating a representation into a linguistic
description and reconstructing or restructuring a representation from a
description. These transitions are not identical to compressing and
unpacking text. Distinctions between processes and structures, and the
connections between them, must be preserved and checked within the
representation, and what is lost, introduced, or changed in the
transition must be identified. The task of language AI is to master this
bidirectional work, not merely to continue text.

If the necessary systemic representations have not been acquired,
understanding requires work to construct them. Such work may take place
in communication, but it cannot be reduced to reading the message and
may take a long time.

This paper's contribution concerns the systemic-activity organization of
the foundations of a message, a separate narrative description of the
situation, and control of transformations between the five levels. The
proposed five levels specify the organization of message formation and
understanding within thinking and activity in which ASI participates.
Its theoretical foundation is considered in Section 2.

Mel'čuk restricts the task of semantic decomposition to ``determining
the applicability of LUs to things and situations already identified''
and excludes the identification of the thing or situation
itself.\footnote{I. Mel'čuk, \emph{Explanatory Combinatorial
  Dictionary}, in Open Problems in Linguistics and Lexicography 225, §
  2.1.1 (G. Sica ed., Polimetrica, Monza, 2006) (pp.~225--355),
  \url{https://olst.ling.umontreal.ca/pdf/Mel\%27c\%CC\%8Cuk_2006_ECD.pdf}.}
This distinction separates linguistic description from the
representations from which a situation is cut out and on whose basis it
is organized.

Reconstructing a situation from a linguistic description already
presupposes schemas through which what is described can be recognized.
Discussing the recognition of another participant in activity, P. G.
Shchedrovitskiy observes: ``And when he encounters the fact that others
are not like him, yet cannot recognize the other person (he has no other
schema, so to speak), a whole host of interpersonal and content-related
conflicts begins''.\footnote{P. G. Shchedrovitskiy, \emph{Введение в
  синтаксис и семантику графического языка СМД-подхода {[}Introduction
  to the Syntax and Semantics of the Graphical Language of the Systemic
  Thought-Activity Approach{]}}, Electronic Library of the G. P.
  Shchedrovitskiy Institute for Development Foundation (Oct.~8, 2010)
  (seventh semester; lecture 38 in the overall cycle; Moscow, Academy of
  National Economy; transcript),
  \url{https://www.fondgp.ru/old/lib/mmk/58/10-10-08_Schedrovitckij_P.G._shematizatciya_-_XXXVIII.doc}.}
What matters for this paper is the connection between recognition and
the availability of a schema. Accordingly, the question of a language
model's reconstruction of a situation first requires establishing which
acquired schemas it uses and which connections it accepts as true when
interpreting the description.

Anna Elashkina explicitly discusses the distinction between ontological
foundations and their textual description in relation to AI: ``Whereas,
previously, such an ontology effectively determined the architecture of
a software system in logical systems, a working ontology now has the
status of a text which, although it may have some priority, remains
merely a text in a neural network and certainly not its
architecture''.\footnote{Anna Elashkina, \emph{Искусственный интеллект
  просит подумать {[}Artificial Intelligence Asks Us to Think{]}},
  Открытые системы. СУБД {[}Open Systems. DBMS{]}, no. 4 (2024),
  \url{https://doi.org/10.51793/OS.2024.81.42.002};
  \url{https://www.osp.ru/os/2024/04/13059028}.} For this paper, the
distinction between functions is essential: a description of an ontology
can be conveyed through context, but the presence of that description is
not identical to the organization of a system in which accepted
distinctions and connections accepted as true serve as foundations for
the formation of a message.

The subject of this paper is the conversational machines of AI systems
that form linguistic messages in joint activity. A person or a
human--machine system participates in constructing this world and in
communication within it. Acquired schemas take shape in a particular
linguistic environment and develop historically.

The acquisition of a representation cannot be reduced to reading or
processing descriptions. A picture of the world is acquired primarily
through actions and activity. N. Ya. Marr connected manual speech with
manual thinking and emphasized its significance ``as a means of
organizing production, behavior, and worldview''.\footnote{N. Ya. Marr,
  \emph{Язык и мышление {[}Language and Thinking{]}}, in 3 Избранные
  работы {[}Selected Works{]} 90, 106 (1934) (pp.~90--121; lecture
  delivered on June 26, 1931; quotation at 106; electronic publication
  by CRECLECO, University of Lausanne),
  \url{https://crecleco.seriot.ch/textes/MARR-31b/txt.html}.} A person
acts, works with material, interacts with other people, and masters
means of activity. Internalization concerns all of this, not only text.
Speech, terminology, meanings, and sense are important, but they do not
exhaust the acquisition of representations.

The paper draws on the author's experience of organizing collective
thinking through the ``Schematization'' technology and on observations
of game participants' speech, schemas, activity, and
communication.\footnote{M. Osovsky, \emph{Технология игры «Схематизация»
  {[}The Technology of the ``Schematization'' Game{]}}, Medium (June 15,
  2018), \url{https://osovsky.medium.com/game-5438c730a15e}.} Between
2011 and 2026, the author conducted more than 350 games involving up to
300 people at a time, using the method of ``organizational activity
games'' described by Marina Liborakina.\footnote{M. Liborakina, \emph{A
  bridge between past and future}, in Games and Simulations to Enhance
  Quality Learning 41 (Simulation and Gaming Yearbook, vol.~4,
  Psychology Press, 1996) (pp.~41--48; ISBN 9780749418663),
  \url{https://books.google.com/books?id=4meAFnexIMoC&dq=Moscow+Methodological+Circle&pg=PA48}.}
Participants jointly describe and schematize the complex systems of
activity in which they are involved, considering their existing
organization in relation to their historical formation and possible
future. In this work, the author observes thinking through the
interrelations of speech, schemas, activity, and communication.
Differences in the identification of elements, connections, and
processes become objects of discussion and of coordination among
participants' representations. This practice provided the grounds for
formulating a task for language AI: the formation of a message must rest
on representations of the activity being described and maintain their
connection to the content being expressed.

The study of thinking in organized joint work has a foundation in the
tradition of organizational activity games. P. G. Shchedrovitskiy
considers such a game a method of comprehensive socio-psychological
research and emphasizes: ``A `game' of this kind is, in essence, real
life activity, but compressed in time and space and taking place under
the artificial conditions of game organization''.\footnote{P. G.
  Shchedrovitskiy, \emph{Организационно-деятельностная игра как метод
  комплексного социально-психологического исследования {[}The
  Organizational Activity Game as a Method of Comprehensive
  Socio-Psychological Research{]}}, in Программно-целевой подход и
  деловые игры {[}The Program-Target Approach and Business Games{]} 36
  (Novosibirsk, 1982) (pp.~36--39; electronic publication),
  \url{https://shchedrovitskiy.com/igra-kak-metod-issledovaniya/}.} The
organization of a game makes it possible to isolate the processes of
thinking and communication occurring within it for investigation. For
this paper, this is important as a basis for observing how an utterance
takes shape within the participants' activity.

Yrjö Engeström develops a related research formulation in the approach
of formative interventions and the practice of the Change Laboratory.
Here, the unit of analysis is a collective activity system. In his study
of work in a Finnish hospital, Engeström shows how participants redefine
the initial problematic situation and the conceptual instrument being
used as they transform their practice.\footnote{Y. Engeström, \emph{From
  Design Experiments to Formative Interventions}, 21 Theory \&
  Psychology 598 (2011) (no. 5; pp.~598--628),
  \url{https://doi.org/10.1177/0959354311419252}.} This approach
provides grounds for considering the development of representations
together with changes in the activity in which they are used.

In \emph{Cognition in the Wild}, Edwin Hutchins investigates the
thinking of a ship's navigation team as the work of a system that
includes participants and the means they use.\footnote{E. Hutchins,
  Cognition in the Wild (MIT Press, Cambridge, MA, 1995) (author's
  introduction), \url{https://pages.ucsd.edu/~ehutchins/citw.html}.} He
traces the propagation and transformation of representations across
different media and treats communication as part of the organization of
collective task performance. What matters for the present formulation is
the shift of the unit of analysis from an isolated individual to joint
activity, within which representations are formed and used.

Kirsikka Vaajakallio and Tuuli Mattelmäki consider collaborative design
games as a means of organizing participants' cooperation.\footnote{K.
  Vaajakallio \& T. Mattelmäki, \emph{Design Games in Codesign: As a
  Tool, a Mindset and a Structure}, 10 CoDesign 63 (2014) (no. 1;
  pp.~63--77; Aalto University research-portal record),
  \url{https://doi.org/10.1080/15710882.2014.881886};
  \url{https://research.aalto.fi/en/publications/design-games-in-codesign-as-a-tool-a-mindset-and-a-structure/}.}
Their work situates the use of games in the research context of jointly
constructing design content. Together with the tradition of
organizational activity games, formative interventions, and research on
distributed cognition, this approach provides a basis for examining
speech and schemas within collective thinking. The author's own
observations and these research approaches provide the grounds for
developing the proposed five-level organization of content.

Comparison with Chomsky clarifies the subject of the proposed
architecture. Chomsky and his coauthors define language as ``a
computational operation occurring in the mind of an individual,
independent of its possible communicative use.''\footnote{J. J. Bolhuis
  et al., \emph{Language: UG or Not to Be, That Is the Question}, 13
  PLOS Biology e1002063 (2015) (no. 2; coauthors: I. Tattersall, N.
  Chomsky, R. C. Berwick),
  \url{https://doi.org/10.1371/journal.pbio.1002063}.} This paper
examines the systemic-activity foundations of message formation:
historically acquired schemas for cutting out a situation,
representations of the world of activity, processes of constructing and
transforming content in communication, and the activity system in which
the means of thinking are formed and used. This distinction concerns
human utterances as well as AI.

Word meanings change over time and differ across communities. The
persistence of a word therefore does not attest to the persistence of
the representations on whose basis it is used.

The author's position is that every language results from the
importation of technologies---situations of activity---together with the
terms that explain them. From this perspective, the transfer of words
must be investigated together with the mastery of that activity and the
transformation of the receiving community's representations. Chomsky, by
contrast, considers language a result of the interaction between the
initial state of the language faculty and experience.\footnote{N.
  Chomsky, New Horizons in the Study of Language and Mind at 4
  (Cambridge University Press, 2000) (ch.~1, p.~4; official excerpt),
  \url{https://doi.org/10.1017/CBO9780511811937};
  \url{https://assets.cambridge.org/97805216/51479/sample/9780521651479wsc00.pdf}.}
What is decisive for this paper is the conceptual representation
(schema) of a specific situation of activity cut out from a systemic
representation of reality as a fragment.

For the argument of this paper, it is important that transferring a term
does not substitute for mastering the situation of activity that it
explains. Knowing a word does not mean acquiring a schema: one must
master the distinctions between processes, structures, and their
connections with which that word is used in the receiving community.

Another problem is that scientific and professional communities assign
their own specialized meanings to the same words. An accepted definition
is incorporated into the norms of professional usage and determines
which assertions and actions a community considers permissible within
its subject domain.

If AI's task is to express a given sense, the representation of that
sense is primary, not the probability of the next word. A probabilistic
generator may select a formulation, but the permissibility of its
content must be determined separately from its linguistic probability.

There are academic grounds for this distinction. In Meaning--Text
theory, a semantic graph represents content that permits different
linguistic realizations.\footnote{I. A. Mel'čuk, \emph{Towards a
  Linguistic ``Meaning ⇔ Text'' Model}, in Trends in Soviet Theoretical
  Linguistics 33 (1973) (pp.~33--57),
  \url{https://doi.org/10.1007/978-94-010-2536-2_2}.} Iordanskaja and
Mel'čuk separately consider the conceptual representation of a situation
from which the semantic representation of a sentence is
constructed.\footnote{L. Iordanskaja \& I. Mel'čuk, \emph{Minimodel of
  Semantic Synthesis of Russian Sentences}, 44 Lingvisticæ
  Investigationes 101 (2021) (no. 1; pp.~101--136),
  \url{https://doi.org/10.1075/li.00058.ior}.} These transitions have
earlier foundations: Kulagina and Mel'čuk distinguish the levels of
situation and sense and discuss the possibility of intermediate levels
between them.\footnote{O. S. Kulagina \& I. A. Mel'čuk, \emph{Automatic
  Translation: Some Theoretical Aspects and the Design of a Translation
  System}, in Machine Translation 137, 145--146 (A. D. Booth ed.,
  North-Holland, Amsterdam, 1967) (pp.~137--171; situation and meaning
  levels, intermediate representations at 145--146),
  \url{https://olst.ling.umontreal.ca/pdf/Mel\%27c\%CC\%8Cuk_1967_Kulagina.pdf}.}
In a later account, Mel'čuk places the Meaning--Text model within a
global model of linguistic behavior that includes extralinguistic
reality and conceptual and semantic representations.\footnote{I.
  Mel'čuk, \emph{Représentation sémantique {[}Semantic
  Representation{]}}, in La théorie Sens-Texte. Concepts-clés et
  applications {[}Meaning--Text Theory: Key Concepts and Applications{]}
  21, 47 (S. Marengo ed., L'Harmattan, Paris, 2021) (pp.~21--47; diagram
  of the global model of linguistic behavior at 47),
  \url{https://olst.ling.umontreal.ca/pdf/Mel\%27c\%CC\%8Cuk_2021_RS\%C3\%A9m.pdf}.}
His 1981 work also considers the semantic representation of an utterance
realized by a sequence of sentences.\footnote{I. A. Mel'čuk,
  \emph{Meaning-Text Models: A Recent Trend in Soviet Linguistics}, 10
  Annual Review of Anthropology 27, 34 (1981) (pp.~27--62),
  \url{https://olst.ling.umontreal.ca/pdf/Mel\%27c\%CC\%8Cuk_MTM_RecentTrends_1981.pdf}.}
Cousin explicitly sets the task of controlling the expression of a
message through formally implemented structures of this
theory.\footnote{M. Cousin, \emph{Dependency Structures Representation:
  Meaning Text Theory's Deep-Syntax Encoding With Abstract Categorial
  Grammars}, MoL 163 (2025) (pp.~163--175),
  \url{https://aclanthology.org/2025.mol-1.13/}.} In research that does
not refer directly to Mel'čuk, Narayan and colleagues introduce a
content plan before generation,\footnote{S. Narayan et al.,
  \emph{Conditional Generation with a Question-Answering Blueprint}, 11
  TACL 974 (2023) (pp.~974--996),
  \url{https://aclanthology.org/2023.tacl-1.55/}.} while the Large
Concept Models team proposes processing at a higher semantic level
followed by decoding into text.\footnote{LCM team, L. Barrault et al.,
  \emph{Large Concept Models: Language Modeling in a Sentence
  Representation Space}, arXiv:2412.08821 (2024),
  \url{https://arxiv.org/abs/2412.08821}.}

The separation of the substantive organization of a message from its
linguistic realization has engineering precedents in classical natural
language generation. Reiter and Dale distinguish document planning,
microplanning, and surface realization. Planning determines the content
and structure of a document, microplanning selects the linguistic means
of expressing it, and surface realization transforms the resulting
specification into text. The communicative goal, the model of the
recipient, and discourse history are taken into account in forming a
message.\footnote{E. Reiter \& R. Dale, Building Natural Language
  Generation Systems chs.~3--6 (Cambridge University Press, 2000)
  (chs.~3--6; official chapter pages),
  \url{https://www.cambridge.org/core/books/abs/building-natural-language-generation-systems/document-planning/6401F096F58AC31DE37A8668BF07BAEF};
  \url{https://www.cambridge.org/core/books/abs/building-natural-language-generation-systems/microplanning/FEB90C3A816AD6A873A1BE85B69A9D0B};
  \url{https://www.cambridge.org/core/books/abs/building-natural-language-generation-systems/surface-realisation/D45A2AC42657177D950874E6322D2609}.}
In the proposed architecture, this separation is supplemented by the
systemic foundations for cutting out a situation and by the distinction
between its description and the full sense of a particular message.

On the side of text comprehension, the model developed by Kintsch and
van Dijk is an important precursor. The authors distinguish the semantic
representation of a text---the textbase---from a model of the situation
to which the text refers. The situation model includes events, actions,
and participants, as well as relevant knowledge and experience that are
not necessarily expressed in the message. They also identify the need
for a model of the communicative context that represents speech acts and
their underlying intentions.\footnote{T. A. van Dijk \& W. Kintsch,
  Strategies of Discourse Comprehension at 12, 338 (Academic Press, New
  York, 1983) (pp.~12, 338; author's copy),
  \url{https://discourses.org/wp-content/uploads/2022/06/Teun-A-van-Dijk-Walter-Kintsch-1983-Strategies-Of-Discourse-Comprehension.pdf}.}
This distinction separates the reconstruction of what was said from the
reconstruction of the situation described and the purpose of the
message. This paper also sets the task of reconstructing the systemic
representations from which that situation was cut out.

In reconstructing G. P. Shchedrovitskiy's position, P. G.
Shchedrovitskiy defines sense as follows:

\begin{quote}
In this position, ``sense'' is defined as the configuration of
connections and relations between different elements of the situation of
activity and the speech process. This configuration is created and
reconstructed by a person understanding the text of a
message.\footnote{P. G. Shchedrovitskiy, \emph{Идея
  «мысли=коммуникации»: смысл vs содержание {[}The Idea of
  ``Thought=Communication'': Sense vs.~Content{]}} §13, Pyotr
  Shchedrovitskiy's Website (Jan.~27, 2025) (last visited Oct.~19,
  2026),
  \url{https://shchedrovitskiy.com/ideya-myisli-kommunikatsii-smyisl-vs-soderzhanie/}.}
\end{quote}

Research on AI also addresses the opacity of the sources of an answer.
Neeman and colleagues show that the entanglement of parametric and
contextual knowledge makes it difficult to establish the source of an
answer and propose generating answers from these sources
separately.\footnote{E. Neeman et al., \emph{DisentQA: Disentangling
  Parametric and Contextual Knowledge with Counterfactual Question
  Answering}, Proceedings of the 61st Annual Meeting of the Association
  for Computational Linguistics (Volume 1: Long Papers) 10056 (2023)
  (pp.~10056--10070; coauthors: R. Aharoni, O. Honovich, L. Choshen, I.
  Szpektor, O. Abend),
  \url{https://doi.org/10.18653/v1/2023.acl-long.559}.}

The political orientation of LLM responses and the absence of political
views made explicit in advance represent another aspect of the opacity
of the foundations of an utterance. Westwood, Grimmer, and Hall show
that users perceive the responses of most of the LLMs studied as
left-leaning, and that an instruction to remain neutral reduces
perceived slant.\footnote{S. J. Westwood et al., \emph{Measuring
  Perceived Slant in Large Language Models Through User Evaluations}
  (2025) (coauthors: J. Grimmer, A. B. Hall),
  \url{https://modelslant.com/paper.pdf}.} Rozado likewise identifies
the political orientation of conversational models' responses and
demonstrates that it can be changed through fine-tuning on politically
oriented data.\footnote{D. Rozado, \emph{The Political Preferences of
  LLMs}, 19 PLOS ONE e0306621 (2024) (no. 7),
  \url{https://doi.org/10.1371/journal.pone.0306621}.} Possible causes
include the effects of pretraining, fine-tuning, evaluators'
preferences, and system instructions. An orientation in the utterance is
observable, but which representations of society, the state, the
economy, and activity underlie it has not yet been established. A change
in perceived slant achieved through an instruction must therefore be
distinguished from the reconstruction and revision of the foundations of
message formation.

The requirement to disclose the foundations of decisions also matters
for AI auditing. Glenn identifies the need for the ``Ability for humans
to check why the AGI is doing something questionable or
undesirable.''\footnote{J. Glenn, Artificial General Intelligence:
  Issues and Opportunities at 9 (Foresight on Demand: Foresight towards
  the 2nd Strategic Plan for Horizon Europe, February 2023) (§ 2.5,
  p.~9),
  \url{https://millennium-project.org/wp-content/uploads/2023/05/EC-AGI-paper.pdf}.}
Applied to a linguistic response, this raises questions not only about
sources of information but also about the value-based foundations for
selecting and organizing content: which representations of society and
activity are accepted, what is considered permissible, desirable, or
impermissible within them, and how these distinctions determine the
message. The proposed architecture sets the task of making such
foundations available for examination, without reducing their
identification to an assessment of the orientation of the finished text.

In the OntoAgent architecture, the ontology, fact repository, and
linguistic lexicon constitute explicitly organized resources for
reasoning, modeling, and communication.\footnote{M. McShane \& S.
  Nirenburg, \emph{A Knowledge Representation Language for Natural
  Language Processing, Simulation and Reasoning}, 6 International
  Journal of Semantic Computing 3 (2012) (no. 1; pp.~3--23),
  \url{https://doi.org/10.1142/S1793351X12400016}.} McShane and Leon
separate the agent's formation of message content from its linguistic
generation.\footnote{M. McShane \& I. Leon, \emph{Language Generation
  for Broad-Coverage, Explainable Cognitive Systems}, arXiv:2201.10422
  (2022), \url{https://arxiv.org/abs/2201.10422}.} Contemporary
developments of this approach envisage using an LLM as an instrument
within a system that retains control over representations and operations
on them. Nirenburg and colleagues describe the acquisition of new
lexical and ontological knowledge with the aid of an LLM whose proposals
are checked against formal ontological constraints.\footnote{S.
  Nirenburg et al., \emph{LLMs as Knowledge Workers}, 8 Proceedings of
  the AAAI Symposium Series 512 (2026) (no. 1; pp.~512--520; coauthors:
  M. McShane, S. Oruganti, J. English, C. Arndt),
  \url{https://doi.org/10.1609/aaaiss.v8i1.42583}.} Oruganti and
colleagues compare OntoAgent with six LLMs within the HARMONIC robotics
system, including cases in which the models receive equivalent
procedural knowledge. In the task investigated, the availability of such
knowledge did not eliminate errors in checking the system's own state of
knowledge, diagnosis, and action selection.\footnote{S. Oruganti et al.,
  \emph{Why Cognitive Robotics Matters: Lessons from OntoAgent and LLM
  Deployment in HARMONIC for Safety-Critical Robot Teaming},
  arXiv:2603.26730 (2026) (coauthors: S. Nirenburg, M. McShane, J.
  English, M. Roberts, C. Arndt, R. Parasuraman, L. Sentis),
  \url{https://arxiv.org/abs/2603.26730}.} These results distinguish the
provision of information to a language model from the organization of a
system in which information is used to form decisions and messages.

In Wikontic, an ``ontology'' is a technical schema of entity types and
permissible connections.\footnote{A. Chepurova et al., \emph{Wikontic:
  Constructing Wikidata-Aligned, Ontology-Aware Knowledge Graphs with
  Large Language Models}, arXiv:2512.00590 (2025) (version 2 of Jan.~29,
  2026; coauthors: A. Bulatov, M. Burtsev, Y. Kuratov),
  \url{https://arxiv.org/html/2512.00590v2}.} In our view, however, it
is neither an ontology in the philosophical sense nor a systemic
representation of activity that distinguishes processes, structures, and
their connections in the sense of this paper.

Nevertheless, working with corpora of texts and their fragments produces
impressive results in language models, although the grounds of this
effectiveness still require explanation. The achievements are
particularly conspicuous in relatively restricted formal languages,
especially programming languages, in which models create and reproduce
code.

The successful solution of a local task may rely on a limited working
ontology of the situation and need not require reconstruction of an
entire system of representations of the world. Understanding the
foundations of the solution requires reconstructing the representations
the model used and the connections between them.

The next task for research and conversational LLMs is to reconstruct the
ontological foundations of linguistic descriptions and to construct
large graphical models of economic organization, the economy,
construction, and software-system development. Reproducing a text about
the architecture of software or an industrial enterprise does not attest
to the reconstruction of that system's schema-model. Such a schema must
preserve the distinctions between processes and structures and the
connections between them and serve as a means of working with them, not
merely as an image.

L. A. Byzov demonstrates the problem of moving from a sequential
description to an integrated representation using historical material:
``Speech therefore inevitably destroys the actual relations between
events occurring simultaneously and at different times (or between
changing states).'' He then observes that ``special mental efforts and
auxiliary means (chronological comparison tables) are needed to
reconstruct from chaos and fragments any reasonably integrated picture
of a historical period''.\footnote{L. A. Byzov, \emph{Докладная записка
  об общих проблемах графического языка и необходимости организации
  Института графического языка {[}Memorandum on the General Problems of
  Graphical Language and the Need to Establish an Institute of Graphical
  Language{]}}, note 71, proposition 6, Electronic Edition of «Мы нашли
  новый язык» {[}``We Have Found a New Language''{]} (note 71: abstracts
  of the paper «История как предмет графического изображения»
  {[}``History as an Object of Graphical Representation''{]},
  proposition 6),
  \url{https://docs.google.com/document/d/170WgJMGu_j_tL4NGumjNcmMZvroC4sTBtNH3h4X-HIQ/edit?tab=t.0}.}
The order of exposition does not coincide with the organization of the
situation described. Understanding requires reconstructing the
connections between events and states within a single representation.
For AI, the crucial question is: which acquired representations and
means does a language model use for such reconstruction? Processing a
sequence of text and producing a coherent response do not, in
themselves, demonstrate that this work has been performed.

In this paper, ASI (superintelligence) means a model capable of
reconstructing the systemic organization of activity: schemas of systems
and subsystems, processes and structures, synchronized timelines of
activity and their connections, and the ontological pictures from which
it acts. Reconstruction means constructing representations with which
the model can work, rather than merely generating their linguistic
descriptions or graphical images. Universality in task performance, the
quality of text, and the ability to modify its own code are not, in
themselves, sufficient indicators of ASI in this sense.

\subsection*{1.1. Token Prediction and the Representation of Sense}
\phantomsection
\label{yale-heading-003}
\addcontentsline{toc}{subsection}{1.1. Token Prediction and the Representation of Sense}

In the GPT approach, next-token prediction is the technical task of
pretraining. In their GPT-2 paper, the authors hypothesized that a
language model with sufficient capacity would begin to learn tasks
represented in natural-language sequences in order to predict those
sequences better.\footnote{A. Radford et al., \emph{Language Models are
  Unsupervised Multitask Learners}, OpenAI (2019) (coauthors: J. Wu, R.
  Child, D. Luan, D. Amodei, I. Sutskever),
  \url{https://cdn.openai.com/better-language-models/language_models_are_unsupervised_multitask_learners.pdf}.}
This hypothesis justifies using language modeling as a means of
acquiring a broad range of capabilities. The GPT-4 technical report
explicitly states that the model is based on the Transformer and
pretrained to predict the next token in a document, after which it is
fine-tuned using human feedback.\footnote{OpenAI, \emph{GPT-4 Technical
  Report}, arXiv:2303.08774 (2023),
  \url{https://arxiv.org/abs/2303.08774}.}

The training task explains the selection of a probable sequence, but the
probabilistic selection of a formulation does not establish the
permissibility of its content. If a language model is given the sense of
a message, it is necessary to check separately whether the selected
formulation expresses precisely that sense and preserves its connections
and conditions. The transition from token prediction to the linguistic
realization of a message therefore requires checking the specified
content, not merely assessing the probability of the text.

Using a probabilistic token generator for the linguistic realization of
a message is, in itself, an extremely controversial experiment with
unpredictable consequences. Its use does not eliminate the need to
organize the foundations of content separately and to check that the
given sense is preserved in the text.

\subsection*{1.2. Context, Internal Memory, and Graph Representations}
\phantomsection
\label{yale-heading-004}
\addcontentsline{toc}{subsection}{1.2. Context, Internal Memory, and Graph Representations}

Developers also use operational definitions of context. OpenAI defines
context as follows: ``Here, context refers to the total window of tokens
(input + output) that the model can attend to at once.''\footnote{E.
  Okcular, \emph{Context Engineering --- Short-Term Memory Management
  with Sessions}, OpenAI Cookbook (Sept.~9, 2025),
  \url{https://developers.openai.com/cookbook/examples/agents_sdk/session_memory}.}
Anthropic defines it as follows: ``Context refers to the set of tokens
included when sampling from a large-language model (LLM).''\footnote{Anthropic,
  \emph{Effective Context Engineering for AI Agents} (Sept.~29, 2025),
  \url{https://www.anthropic.com/engineering/effective-context-engineering-for-ai-agents}.}
When discussing external notes, the authors specify: ``These notes get
pulled back into the context window at later times.''\footnote{Anthropic,
  \emph{Effective Context Engineering}, \emph{supra} note 38, §
  Structured note-taking.}

An agent system selects instructions, message history, materials, and
tool results for its next call to the model.\footnote{Anthropic,
  \emph{Effective Context Engineering}, \emph{supra} note 38.}
Distributed vector representations emerge during context processing. To
accelerate generation, a KV cache retains the attention keys and values
computed for positions already processed, allowing them to be reused
without recomputation.\footnote{Hugging Face, \emph{How caching works},
  Transformers Documentation (last visited Oct.~18, 2026),
  \url{https://huggingface.co/docs/transformers/en/cache_explanation}.}
The cache is a numerical state of computation, not an explicitly labeled
inventory of participants, predicates, conditions, and unknowns.

The Claude Code documentation describes the agent's work as a recurring
cycle of three phases: gathering context, taking action, and verifying
results. The results of tool use return to the model and influence its
selection of subsequent actions.\footnote{Anthropic, \emph{How Claude
  Code works}, Claude Code Documentation,
  \url{https://code.claude.com/docs/en/how-claude-code-works}.} Such a
cycle organizes the model's interaction with its environment. The five
levels proposed in this paper address a different task: distinguishing
systemic foundations, the conceptual representation of a situation, its
narrative description, the full sense of a message, and text. Gathering
context, acting, and checking results do not, in themselves, determine
the systemic representations from which a situation has been cut out or
how its content has been transformed into a message. The agent cycle and
the five-level organization of content describe different aspects of
AI's work: the former describes the sequence of interaction with the
environment, whereas the latter describes the organization of
representations and transitions between them.

In the xAI materials examined, the emphasis is on knowledge retrieval
and context length. The announcement of Grok states: ``We are working on
methods that can discover and retrieve information whenever it is
needed.''\footnote{xAI, \emph{Announcing Grok} (Nov.~3, 2023),
  \url{https://x.ai/blog/grok}.} In the announcement of Grok-1.5,
increased ``memory capacity'' is explicitly linked to expanded context
length: ``an increased memory capacity of up to 16 times the previous
context length.''\footnote{xAI, \emph{Announcing Grok-1.5} (Mar.~28,
  2024), \url{https://x.ai/news/grok-1.5}.}

An agent's long-term memory may store histories, summaries, plans, and
results in external files or databases. Compression of materials and
structured notes help it continue long-running tasks.\footnote{Anthropic,
  \emph{Effective Context Engineering}, \emph{supra} note 38.}

In \emph{Memorizing Transformers}, the authors propose a different
memory mechanism: ``we extend language models with the ability to
memorize the internal representations of past inputs.''\footnote{Y. Wu
  et al., \emph{Memorizing Transformers}, abstract, ICLR (2022)
  (coauthors: M. N. Rabe, D. Hutchins, C. Szegedy),
  \url{https://arxiv.org/abs/2203.08913}.} Additional retrievable memory
differs from knowledge in the weights and from the current context
window.

When selecting, compressing, and reusing materials, it is necessary to
preserve the level of representation to which they belong, their
connections and constraints, and the grounds of the information
accepted.

Research by Google Research and DeepMind distinguishes context from
parametric knowledge. In their study of T5 and PaLM, Li and colleagues
formulate a hierarchy of preferred sources: ``relevant context
\textgreater{} model's parametric knowledge \textgreater{} irrelevant
context'' (equations (1)--(2)).\footnote{D. Li et al., \emph{Large
  Language Models with Controllable Working Memory}, Findings of the
  Association for Computational Linguistics: ACL 2023, 1774, 1775 (2023)
  (pp.~1774--1793; coauthors: A. S. Rawat, M. Zaheer, X. Wang, M.
  Lukasik, A. Veit, F. Yu, S. Kumar),
  \url{https://aclanthology.org/2023.findings-acl.112/}.} In
\emph{Racing Thoughts}, the authors observe: ``Out-of-order
contextualization can result in incorrect or inconsistent LLM
generations.''\footnote{M. A. Lepori et al., \emph{Racing Thoughts:
  Explaining Contextualization Errors in Large Language Models}, NAACL
  3020 §1 (2025) (pp.~3020--3036; coauthors: M. C. Mozer, A.
  Ghandeharioun), \url{https://aclanthology.org/2025.naacl-long.155/}.}

OpenAI research considers acquired internal representations separately
from the current input material. Gao and colleagues write: ``Since
language models learn many concepts, autoencoders need to be very large
to recover all relevant features.''\footnote{L. Gao et al.,
  \emph{Scaling and Evaluating Sparse Autoencoders}, abstract, OpenAI,
  arXiv:2406.04093 (2024), \url{https://arxiv.org/abs/2406.04093}.}

The presence of trained LLM weights does not, in itself, attest to the
presence of deliberately constructed systemic representations of the
world of activity. This requires establishing whether the model
distinguishes processes, structures, and their connections and uses
these distinctions to cut out a situation and form and understand a
message.

The opacity of content selection also appears in the generation of
fictional stories. In \emph{Elias in the Lighthouse, Again?}, 88.3\% of
20,000 stories generated by four language models contained at least one
of 11 persistently recurring words denoting names, occupations, and
settings. Yet stories containing such words accounted for only about
3.8\% of stories in the OLMo fine-tuning data examined. The frequency of
these elements in the corpus does not explain their predominance in
responses as a simple reproduction of the distribution of training
examples.\footnote{S. Hamilton \& D. Mimno, \emph{Elias in the
  Lighthouse, Again? Diagnosing Low Diversity in LLM Stories},
  arXiv:2605.26492 (2026) (version 1),
  \url{https://arxiv.org/html/2605.26492v1}.} A user's request for a
story does not specify particular characters and situations, yet models
systematically introduce similar content. The existence of a trained
mechanism and an available corpus therefore does not yet reveal the
foundations on which a language model cuts out a situation and selects
the content of a message.

A computational graph describes dependencies between operations,
attention connections describe interactions between positions during
processing, and a semantic graph describes the sense-related connections
of a linguistic description. Attribution graphs reconstruct the
influence of internal features through an approximate replacement
model.\footnote{E. Ameisen et al., \emph{Circuit Tracing: Revealing
  Computational Graphs in Language Models}, Anthropic (Mar.~27, 2025),
  \url{https://transformer-circuits.pub/2025/attribution-graphs/methods.html}.}

In an Anthropic study, Templeton and colleagues observe: ``Many features
are multilingual (responding to the same concept across languages) and
multimodal (responding to the same concept in both text and
images).''\footnote{A. Templeton et al., \emph{Scaling Monosemanticity:
  Extracting Interpretable Features from Claude 3 Sonnet}, introductory
  part, Anthropic (May 21, 2024),
  \url{https://transformer-circuits.pub/2024/scaling-monosemanticity/index.html}.}
In their study of Claude 3.5 Haiku, the authors write: ``We find the
model uses a mixture of language-specific and abstract,
language-independent circuits.''\footnote{Anthropic, \emph{On the
  Biology of a Large Language Model}, introduction (2025),
  \url{https://transformer-circuits.pub/2025/attribution-graphs/biology.html}.}

A 2026 study by researchers from Google DeepMind, Google Research, and
universities distinguishes the encoding of novel semantics from their
use in open-weight models: ``open-weights LMs struggle to deploy
representations of novel semantics that are defined in-context, even if
they encode these semantics in their latent
representations.''\footnote{M. A. Lepori et al., \emph{Language Models
  Struggle to Use Representations Learned In-Context}, abstract,
  arXiv:2602.04212 (2026) (version 2; coauthors: T. Linzen, A. Yuan, K.
  Filippova), \url{https://arxiv.org/html/2602.04212v2}.} In the
topology experiment, the authors report: ``LMs achieve substantially
higher accuracy when given the explicit description than when they need
to infer the topology in-context.''\footnote{Lepori et al.,
  \emph{Language Models Struggle to Use Representations Learned
  In-Context}, \emph{supra} note 54, §4.} These studies distinguish the
presence of material, its internal representation, and its functional
use.

The recent ThoughtComm study investigates communication between AI
agents through models' internal representations. Contextualized agent
states are transformed into a shared latent representation from which
individual continuous prefixes are formed for subsequent generation. A
channel for transmitting internal representations thus supplements
textual exchange.\footnote{Y. Zheng et al., \emph{Thought Communication
  in Multiagent Collaboration}, 38 Advances in Neural Information
  Processing Systems (2025) (coauthors: Z. Zhao, Z. Li, Y. Xie, M. Gao,
  L. Zhang, K. Zhang),
  \url{https://proceedings.neurips.cc/paper_files/paper/2025/hash/b2b502c3629beadda06311386d2c6f73-Abstract-Conference.html}.}
The question of transmitting such representations must be distinguished
from that of the organization of the object of inquiry they represent:
which processes, structures, and situations they represent, and on what
foundations these have been identified.

Taken together, these sources distinguish contextual tokens, parametric
knowledge, additional memory mechanisms, and internal conceptual
representations. Investigating internal mechanisms does not replace
establishing the foundation that determined the formation of a
particular message.

Maintaining the content of a whole text has a separate foundation in P.
G. Shchedrovitskiy's text:

\begin{quote}
Developing what the speaker intends to express and planning the speech,
selecting individual sentences and words, changing the order of
exposition, and, for more complex texts, creating a system of references
and footnotes indicate that, even when an individual sentence is being
written down or spoken, both the preceding and the following fragments
of the text must be maintained in memory in accordance with its
subject-matter content.\footnote{Shchedrovitskiy, \emph{Идея
  «мысли=коммуникации» {[}The Idea of ``Thought=Communication''{]}},
  \emph{supra} note 24, §17.}
\end{quote}

This requirement to maintain the subject-matter content of a whole text
provides a transition to distinguishing the levels of its organization
and transformation into a message.

\section*{2. Five Levels of Organization}
\phantomsection
\label{yale-heading-005}
\addcontentsline{toc}{section}{2. Five Levels of Organization}

The author proposes five levels of content organization on the basis of
experience organizing collective thinking in the ``Schematization''
game. During the game, participants jointly construct and discuss
representations of complex activity systems. An individual participant
proceeds from acquired systemic representations of the world of
activity, describes them in a graphical language, cuts out a particular
situation from those representations as a fragment, elaborates its
schema into a graphical description, determines together with other
participants the content needed for a message, and expresses it in
graphical schemas. Dialogue accompanies this communication at a fairly
primitive level. Indeed, very primitive dialogue messages devoid of
substantive content can support the participants' actual communication.

When understanding another participant's message, the work proceeds in
the reverse direction: from the graphical schemas presented to
reconstruction of the message's content, the extended description of the
situation, its conceptual representation, and the systemic
representations from which that situation was cut out. The proposed
architecture transfers this organization of message formation and
understanding to language-based ASI by distinguishing systemic
representations, the conceptual representation of a situation, its
narrative description, the full sense of a message, and text. What is
transferred comprises the distinctions within content and the directions
of work on it, not the assignment of graphical expression to a single
level: graphical schemas are available at each of the five levels, not
only at the level of situation representation.

In the linguistic architecture, an extended graphical description of a
situation is transferred into a sequential verbal description at the
third level. What is transferred is the function of content
organization---the elaboration and comparison of the representations of
the processes, structures, and connections being described---not the
graphical means itself. Graphical schemas remain means of work at every
level. The third level describes the situation, whereas the fourth
selects and organizes from that description the content required for a
particular message in light of the communicative goal.

\textbf{The first level: deliberately constructed systemic
representations of the world of activity.} Communication in joint
activity requires distinguishing the processes taking place, the
structures in which they occur, and their connections. Without these
distinctions, it is impossible to establish which organization of
activity has been accepted in cutting out a situation and selecting a
message. The representations use a graphical language that distinguishes
processes, functional and morphological structures, connections,
material, price, value, and evaluation, as well as past, present, and
future. This systemic dimension does not arise by adding nodes to a
semantic network. Graphical signs have a specified interpretation and
serve as means of working with a representation.

\textbf{The second level: the conceptual representation of a particular
situation.} A situation is cut out from first-level representations as a
fragment in light of the position of a participant in joint activity.
Its schema-model preserves the organization of that fragment---its
elements, structures, connections, processes, functions, evolution, and
dynamics. Mel'čuk relates the transition from extralinguistic reality to
linguistic sense to a representation of the world as perceived by the
speaker.\footnote{I. Mel'čuk, Vers une linguistique Sens-Texte {[}Toward
  a Meaning--Text Linguistics{]} (Collège de France, 1997) (inaugural
  lecture),
  \url{https://olst.ling.umontreal.ca/pdf/Mel\%27c\%CC\%8Cuk_1997_Coll\%C3\%A8ge_de_France.pdf}.}

\textbf{The third level: a semantic description of the situation in
narrative form.} The story elaborates the selected description
sequentially through the community's natural language. A narrative
elaborates content sequentially through linguistic schemas; a graphical
schema allows representations of processes, structures, and their
connections to be maintained and compared jointly and simultaneously.
The description of participants, connections, properties, and conditions
must preserve the content of the situation. A story differs from what
needs to be conveyed in a particular communication.

\textbf{The fourth level: the full sense of the message.} Communicative
goals guide the selection and organization of content. The full sense
represents the content of a particular message about a situation; the
speaker's intention is not included in it. Essential conditions,
negation, and what remains unknown are maintained here. Conveying
everything necessary for a particular message does not mean telling
everything about the situation. This level is a representation of sense,
not the text that expresses it.

\textbf{The fifth level: text, the linguistic realization of the
message.} The language model selects words, syntax, and the order of
exposition. Abbreviation is permissible if the recipient can reconstruct
the full sense at the fourth level through jointly constructed and
agreed-upon representations, the situation of activity, and previous
communication. It must not introduce new assertions or change
connections, conditions, or what remains unknown.

In the reverse direction, the full sense of the message is reconstructed
first, followed by an extended description of the situation, then its
schema-model and the systemic foundations from which it was cut out.
Reconstruction does not mean reconstructing the complete original
situation from a limited message: the work concerns the connections
essential to understanding, not every detail of the situation.
Connections that are neither expressed in the message nor confirmed in
joint activity remain unknown and must not be filled in as established
facts. The extended description makes it possible to identify which
connections are necessary for understanding and which still require
clarification before being incorporated into the situation schema and
related to systemic foundations.

\begin{samepage}
The two directions of work across the five levels are shown in the
diagram (Figure 1).
\par\end{samepage}

\begin{figure}[H]
\centering
\includegraphics[width=\linewidth,keepaspectratio]{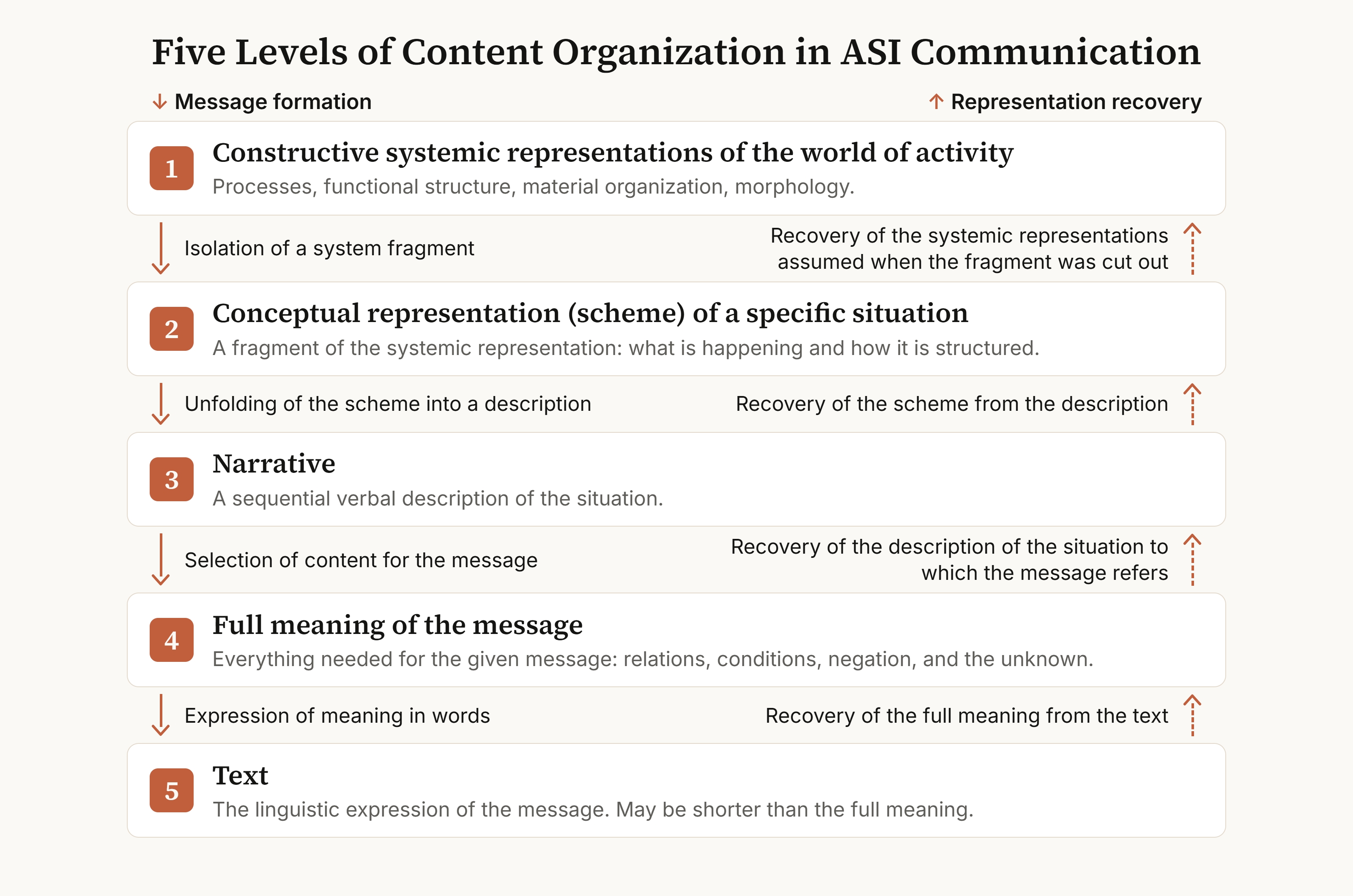}
\par\vspace{6pt}{\fontsize{10pt}{12pt}\selectfont Figure 1. Five Levels of Content Organization in ASI Communication\par}
\end{figure}

The incompleteness of a systemic representation is consistent with A.
Korzybski's structural differential: a perceived object is an
abstraction from a process, and its verbal description is a further
abstraction. At each transition, some characteristics remain outside the
resulting representation. Korzybski emphasizes that ``the object is not
the event {[}\ldots{]} and that the label is not the object nor the
event.'' A description must therefore be distinguished from what it
describes.\footnote{A. Korzybski, Science and Sanity: An Introduction to
  Non-Aristotelian Systems and General Semantics at 389 (1933) (ch.~XXV:
  On the Structural Differential; pp.~386--411; quotation at 389),
  \url{http://esgs.free.fr/uk/art/sands-ch25.pdf}.}

G. P. Shchedrovitskiy provides a theoretical foundation for systemic
representation. In ``Two Concepts of System'' (1974), he writes:

\begin{quote}
``To consider an object as a complex system means to represent it
successively in four categorial planes: processes of some one kind,
functional structure, organized forms of material,
morphology\ldots''.\footnote{G. P. Shchedrovitskiy, \emph{Два понятия
  системы {[}Two Concepts of System{]}}, in Избранные труды {[}Selected
  Writings{]} (1974) (electronic publication),
  \url{https://gtmarket.ru/library/basis/3961/3972}.}
\end{quote}

In the later account, ``Methodology and Philosophy of Organizational and
Managerial Activity,'' G. P. Shchedrovitskiy distinguishes five planes:
processes, functional structures, structures of connections, organized
forms of material, or morphology, and material. He explains the
transition from four to five by separating functional structures from
structures of connections.\footnote{G. P. Shchedrovitskiy, Методология и
  философия организационно-управленческой деятельности {[}Methodology
  and Philosophy of Organizational and Managerial Activity{]} (From the
  Archive of G. P. Shchedrovitskiy, vol.~5, 2003) (quotation
  republication),
  \url{https://priss-laboratory.net.ru/S.C.H.E.M.E.S.-/museum-schemes/museum-schemes_2.htm}.}
This passage is reproduced in A. P. Zinchenko's reader.\footnote{A. P.
  Zinchenko, Путеводитель по основным понятиям и схемам методологии
  Организации, Руководства и Управления {[}A Guide to the Basic Concepts
  and Schemas of the Methodology of Organization, Leadership, and
  Management{]}, at PDF p.~76 (Delo, Moscow, 2004) (G. P.
  Shchedrovitskiy excerpt at PDF p.~76),
  \url{https://pqm-online.com/assets/files/lib/books/zinchenko.pdf}.}
``Programming Scientific Research and Development'' likewise formulates
the requirement to stratify an object into ``at least five planes of
existence''.\footnote{G. P. Shchedrovitskiy, Программирование научных
  исследований и разработок {[}Programming Scientific Research and
  Development{]} §2.7.2 (From the Archive of G. P. Shchedrovitskiy,
  vol.~1, 1999) (pt.~I, ch.~2, §2.7.2),
  \url{https://gtmarket.ru/library/basis/6473/6476}.}

The systemic planes specify the organization of first-level
representations.

Examining foundations means establishing which schemas and systemic
representations have been adopted and which connections have been
accepted as true in forming a message. Adjustment means explicitly
changing them, including revising distinctions and connections. This
work differs from selecting additional texts for the context.

Scientific facts and definitions are not ready-made givens independent
of ways of constructing the object of inquiry, conducting research, and
securing collective recognition. A representation of a situation must
therefore distinguish what is established from what is assumed and what
remains unknown, and disclose the grounds of that distinction. Ludwik
Fleck's concept of a thought style addresses the historical and
collective conditioning of facts and definitions.\footnote{W. Sady,
  \emph{Ludwik Fleck}, Stanford Encyclopedia of Philosophy (substantive
  revision: Dec.~18, 2025) (§§ 5--6: the formation of facts and
  definitions within a thought style),
  \url{https://plato.stanford.edu/entries/fleck/}.}

Mel'čuk's full semantic representation, SemR, provides support for the
fourth level. In his 2014 work, Mel'čuk describes the SemR of a sentence
as a combination of semantic, semantic-communicative, rhetorical, and
referential structures.\footnote{I. Mel'čuk, \emph{Dependency in
  Language}, in Dependency Linguistics: Recent Advances in Linguistic
  Theory Using Dependency Structures at 4--5 (K. Gerdes, E. Hajičová \&
  L. Wanner eds., John Benjamins, 2014) (components of the full SemR at
  4--5),
  \url{https://olst.ling.umontreal.ca/pdf/Mel\%27c\%CC\%8Cuk_2014_Dependency_in_Language.pdf}.}
An earlier work already considers the SemR of an utterance realized by a
sequence of sentences and the construction of semantic subnetworks for
individual sentences.\footnote{Mel'čuk, \emph{Meaning-Text Models},
  \emph{supra} note 18, at 34, 44.} The present architecture uses this
four-component characterization in organizing the sense of a message in
joint activity and defines its place in relation to narrative and the
recipient's understanding.

Moreover, \textbf{words do not have definitive meanings established
independently of human history and activity.} Language is not only used
to express content that has already been constructed; it is also
continually transformed in the development of new representations.

On the question of the definiteness of linguistic meaning, this paper
adopts a polemical position toward Mel'čuk. Kulagina and Mel'čuk
characterize two expressions as having ``the same and quite definite
meaning,'' which can refer to different situations.\footnote{Kulagina \&
  Mel'čuk, \emph{Automatic Translation}, \emph{supra} note 16, at 142.}
The author rejects treating the definiteness of meaning as an initial
characteristic of an expression: in the proposed formulation, its
foundation must be established through acquired representations and the
practice of use in a particular situation of activity. Stable usage is
possible, but it is subject to revision when these foundations change.

The very organization of a picture of the world undergoes historical
transformation, not merely the stock of information within it. The
foundations for identifying objects, determining their properties, and
establishing connections change. The persistence of individual terms and
results does not attest to the invariance of the whole. An explicit
systemic foundation is therefore not a definitive universal ontology: it
must be available for examination and revision.

In P. G. Shchedrovitskiy's text ``The Idea of `Thought=Communication':
Sense vs.~Content,'' schemas are considered means of organizing
thinking:

\begin{quote}
The systemic thought-activity approach assumes that these units are
``schemas,'' and that the key process is ``schematization''.\footnote{Shchedrovitskiy,
  \emph{Идея «мысли=коммуникации» {[}The Idea of
  ``Thought=Communication''{]}}, \emph{supra} note 24, §20.}
\end{quote}

The norms of language and thinking are explicitly distinguished:

\begin{quote}
The fact that the rules (norms) of ``logic'' and ``grammar,'' and, more
broadly, the norms of ``language'' and the norms of ``thinking,'' do not
coincide was discovered a very long time ago.\footnote{Shchedrovitskiy,
  \emph{Идея «мысли=коммуникации» {[}The Idea of
  ``Thought=Communication''{]}}, \emph{supra} note 24, §18.}
\end{quote}

The source itself limits the application of the idea of
thought-communication:

\begin{quote}
Thought=communication takes shape only within complete systems of
thought-activity.\footnote{Shchedrovitskiy, \emph{Идея
  «мысли=коммуникации» {[}The Idea of ``Thought=Communication''{]}},
  \emph{supra} note 24, §23.}
\end{quote}

The disagreement with Chomsky also concerns the role of language in
constructing thought. Chomsky and his coauthors associate the operation
Merge with the construction of hierarchically organized objects of the
``language of thought.''\footnote{J. J. Bolhuis et al., \emph{How Could
  Language Have Evolved?}, 12 PLOS Biology e1001934 (2014) (no. 8;
  coauthors: I. Tattersall, N. Chomsky, R. C. Berwick),
  \url{https://doi.org/10.1371/journal.pbio.1001934}.} In the proposed
architecture, content organization is not derived from linguistic
computation: cutting out a situation and forming a message rely on
systemic representations of the world of activity and connections
accepted as true. The question is therefore not only how an expression
is constructed, but also on what foundations the processes, structures,
and connections described are identified. Language's participation in
the development of representations is acknowledged, but this does not
make the linguistic mechanism a sufficient foundation of content.

Mel'čuk and Polguère write: ``we are looking for the semantic primes of
individual languages and not for semantic primes of `human
thought.'\,''\footnote{I. Mel'čuk \& A. Polguère, \emph{Theory and
  Practice of Lexicographic Definition}, 19 Journal of Cognitive Science
  417, 460--461 (2018) (no. 4; pp.~417--470),
  \url{https://olst.ling.umontreal.ca/pdf/Mel\%27c\%CC\%8Cuk_Polgu\%C3\%A8re_2018.pdf}.}
The systemic categories of the present architecture organize
representations of the world of activity, whereas linguistic semantic
units organize the sense of its description.

Systemic representations are acquired and transformed in communication
connected with activity. A participant cuts out a particular situation
from these representations and determines the content of a message in
joint work. This is the common point of disagreement with the positions
of Mel'čuk and Chomsky considered here. Chomsky's description of an
internal linguistic mechanism and Mel'čuk's account of the transition
from a situation through conceptual representation to sense and text do
not disclose this organization of the acquisition and use of systemic
representations.

At the same time, \textbf{different scientific schools may divide the
same domain of reality differently and construct different objects of
inquiry within it.} Comparing schools requires reconstructing the
accepted principles of identifying objects and connections, including
distinctions for which another system has no direct equivalent. Kulagina
and Mel'čuk distinguish a common situation being described from identity
of linguistic meaning.\footnote{Kulagina \& Mel'čuk, \emph{Automatic
  Translation}, \emph{supra} note 16.} The present architecture locates
the comparison of ways of constructing an object of inquiry in work on
systemic foundations, not merely on linguistic semantics.

On the one hand, \textbf{translation is impossible as a guaranteed
substitution of words that preserves content independently of the
foundations of the original message.} Its task is to reconstruct the
essential content of the original description, the expression of which
in another language may require creating new linguistic means.

On the other hand, \textbf{translation is possible only insofar as
participants can establish correspondences between working
representations in a particular situation of activity.} Coordinating the
limited working ontology necessary for joint activity, rather than an
entire picture of the world, is sufficient for this purpose. A shared
vocabulary serves as a means of such coordination and is developed or
refined in its course.

Kulagina and Mel'čuk provide situational support for this proposition.
Considering expressions with different linguistic meanings, they write:
``they turned out to be translation (not meaning!) equivalents owing
only to the fact that their different meanings can describe the same
situation.''\footnote{Kulagina \& Mel'čuk, \emph{Automatic Translation},
  \emph{supra} note 16, at 145.}

\section*{3. Architectural Implications}
\phantomsection
\label{yale-heading-006}
\addcontentsline{toc}{section}{3. Architectural Implications}

The proposed organization shifts control of a linguistic response from
the finished text to transformations of content. For each transition,
the input information, permissible operations, and criteria for
preserving content must be specified. Missing information may be
obtained from a new source, inferred through an explicitly stated rule,
or clarified with the recipient.

Changing a foundation differs from clarifying a situation and changing a
message. Revising an acquired schema, a systemic distinction, or a
connection accepted as true requires checking whether the situation cut
out and the content based on it remain valid. Clarifying information or
the boundaries of a situation may require revising the narrative and the
sense of the message without changing the first level. A change in the
communicative goal changes the selection of content, whereas
reformulation concerns its linguistic realization. In each case, the
representations affected and the consequences of the change must be
established in both directions: for the formation of a description and
message, and for the reverse reconstruction of the situation and its
systemic foundations.

Control checks whether a process has been replaced by its result,
whether structure, action, and function have been conflated, and whether
roles have been attributed to participants or structures without
grounds.

Control operates both in the movement toward text and in the movement
from text to foundations. In the forward direction, it checks the
preservation of accepted content during elaboration, selection, and
linguistic realization. In the reverse direction, it checks the
foundations for reconstructing the full sense, extended description, and
situation schema, and identifies the systemic distinctions used. A
discrepancy may require clarification of the text or sense, revision of
the description and of how the situation is cut out, or a return to
systemic foundations. The location at which an error is detected does
not limit the direction in which it can be corrected.

Separating the levels makes it possible to localize errors: a categorial
error at the first level, cutting out a wrong fragment or element at the
second, distortion of the story at the third, erroneous selection or
organization of a message at the fourth, and ambiguous formulation or
unjustified abbreviation at the fifth.

Weaver distinguishes the technical accuracy of symbol transmission, the
accuracy of conveying the intended sense, and the effect of the received
sense on the recipient's behavior.\footnote{W. Weaver, \emph{Recent
  Contributions to the Mathematical Theory of Communication}, in C. E.
  Shannon \& W. Weaver, The Mathematical Theory of Communication at 4
  (1949) (p.~4; scan of reprint, tenth printing, 1964),
  \url{https://monoskop.org/images/b/be/Shannon_Claude_E_Weaver_Warren_The_Mathematical_Theory_of_Communication_1963.pdf}.}
These questions concern different aspects of a single act of
communication. Evaluating language-based ASI requires distinguishing the
correctness of message reproduction, the reconstruction of its content
and foundations, and the use of what has been understood in joint
activity.

The preservation of a message's sense must be considered in light of the
position of the person understanding it. The source describes this as
follows:

\begin{quote}
In developed forms, highlighting such relations of sense presupposes the
``speaker's'' adoption of the position of the ``person understanding,''

simulation of the processes of ``understanding'' and
``non-understanding,'' and the introduction into the initial text of
additional constructions oriented toward dialogue.\footnote{Shchedrovitskiy,
  \emph{Идея «мысли=коммуникации» {[}The Idea of
  ``Thought=Communication''{]}}, \emph{supra} note 24, §21.}
\end{quote}

\subsection*{3.1. Understanding a Message and the Criterion of ASI Adequacy}
\phantomsection
\label{yale-heading-007}
\addcontentsline{toc}{subsection}{3.1. Understanding a Message and the Criterion of ASI Adequacy}

Language and understanding are different processes: knowing words does
not guarantee understanding, and the absence of a shared language does
not make understanding impossible. P. G. Shchedrovitskiy shows that
verbal meanings alone are insufficient for the formation of sense: ``But
imagine that introducing meaning does not help: all the words have been
explained, yet it is still not clear; no sense is formed. Then another
function must be brought into play: knowledge must be drawn
upon''.\footnote{P. G. Shchedrovitskiy, \emph{Выготский Л. С. и
  современная педагогическая антропология {[}L. S. Vygotsky and
  Contemporary Pedagogical Anthropology{]}}, in Тьюторство: идея и
  идеология : материалы 1-й Межрегиональной тьюторской конференции,
  Томск, 29 февраля --- 2 марта 1996 г. {[}Tutoring: Idea and Ideology:
  Proceedings of the First Interregional Tutoring Conference, Tomsk,
  February 29--March 2, 1996{]} 73 (I. D. Proskurovskaya ed., School of
  Cultural Policy, Tomsk, 1996) (pp.~73--128; third lecture, section
  «Знаки, смыслы, значения и знания как организованности
  мыследеятельности» {[}``Signs, Senses, Meanings, and Knowledge as
  Organized Forms of Thought-Activity''{]}),
  \url{https://shchedrovitskiy.com/vygotskij-i-sovremennaja-pedagogicheskaja-antropologija}.}
This distinction also matters for evaluating language AI: explaining the
words used does not yet attest to reconstruction of the situation and
understanding of the message. In a limited situation, participants'
actions may serve as means of establishing sense without a verbal
message. Discussing an ambiguous text, Kulagina and Mel'čuk write that
``we cannot rely on our knowledge of the respective language'' and that
``what we need is knowledge of corresponding reality.''\footnote{Kulagina
  \& Mel'čuk, \emph{Automatic Translation}, \emph{supra} note 16, at
  144.} P. G. Shchedrovitskiy distinguishes the positions of the speaker
and the person understanding.\footnote{Shchedrovitskiy, \emph{Идея
  «мысли=коммуникации» {[}The Idea of ``Thought=Communication''{]}},
  \emph{supra} note 24, §4.} A discrepancy requires establishing which
representations must be revised and how this will change the content of
the message.

Recognizing words and the verbal form of a request does not mean
understanding the task. Without a representation of activity, a
participant may consider the message understandable yet incorrectly
identify the required action.

Maturity here is defined by mastery of the means of organizing thinking:
participants construct systemic representations of activity and use
schemas to coordinate what must be communicated and understood. The
completeness of such coordination concerns the essential content of the
message, not all the speaker's individual experiences.

In organizing collective thinking, the author observes a distinction
between constructing a representation of activity and producing expected
utterances about it. A participant may command the terminology,
reproduce accepted explanations, and consistently connect phrases that
fit one another while failing to maintain within a representation the
distinctions between processes and structures and the connections of the
activity described. In such cases, verbal coherence does not express
constructed subject-matter content but substitutes for the work of
constructing it. The continuation of speech is determined by which
phrase fits the preceding one and is expected by other participants,
rather than by which established connection must be expressed.

When expected phrases are being produced, the organization of the
activity described is not presented for examination and verification. A
clarifying question may elicit a new verbal explanation that maintains
the coherence of the conversation without resolving the discrepancy
discovered in the organization of the object of inquiry. The mere fact
that a conversation is maintained does not mean that participants have
common representations of activity. Their joint construction is a
separate undertaking, not an automatic result of exchanging messages.

The requirement to represent processes and structures graphically
reveals the difference between these ways of working. A participant must
separate process from structure, show connections between the organized
forms identified, relate the sequence of events to the simultaneously
existing organization of activity, and preserve accepted distinctions
when considering different fragments. What is unknown must be marked as
unknown, rather than concealed by a familiar formulation. As
participants jointly construct and agree upon a schema, different
representations of the organization of activity are presented together,
and the discrepancies discovered become objects of discussion and
transformation of representations. In communication connected with
activity, participants acquire and transform their own and others'
representations, as well as jointly constructed and agreed-upon
representations through questions, comparison of explanations,
identification of discrepancies, and modification of schemas.

Speech substitutes for the construction of content when further speaking
is directed toward maintaining the persuasiveness of an answer and a
contradiction discovered is covered over with another expected phrase.

The observed similarity between this kind of human speaking and
language-model generation concerns the way the work is performed.
Producing an expected verbal continuation substitutes for constructing
and checking the systemic representations that provide foundations for
both an utterance about activity and the organization of the activity
itself. Reproducing ordinary human speech therefore does not provide a
sufficient criterion for strong language AI. The organization of content
formation must be deliberately developed in both humans and AI:
establishing the foundations for cutting out a situation, constructing
representations, checking their transformations, and using the results
of those checks when forming a message.

The criterion is the ability to move from what has been said to a
representation of the object of inquiry, work with it, and change the
message after revising that representation. It is precisely this ability
that makes it possible to \textbf{distinguish verbal success from
technologically organized thinking} in joint activity.

Boroday maintains that ``any correlation of a sign with subject-matter
reality takes place through the prism of mental content; that is,
reference is impossible without conceptualization''.\footnote{S. Yu.
  Boroday, Язык и познание: Введение в пострелятивизм {[}Language and
  Cognition: An Introduction to Postrelativism{]} at 583 (V. A.
  Lektorsky \& A. V. Smirnov eds., Sadra LLC: YaSK Publishing House,
  Moscow, 2020) (800 pp.; ISBN 978-5-906859-62-4; propositions cited at
  583--584, 593).} In this formulation, the foundation of the
correlation is located in the individual psyche: a sign is correlated
with the content of the individual's conceptual system, through which
subject-matter reality is recognized. Although Boroday takes into
account the social origins of linguistic conventions and the influence
of cultural practices, the immediate explanation of reference proceeds
through individual mental content.

In the author's view, the foundation for correlating a sign with reality
should be sought in a situation of activity. A participant uses a sign
while occupying a particular position in joint work and correlates it
with what has been identified as the object of that work. Establishing
what an utterance refers to therefore means reconstructing the situation
within which it is produced and used. A participant's mental content may
accompany this process, but it does not, in itself, determine which
correlation is accepted in the activity or on what foundations other
participants can understand and check it. A graphical schema serves here
as a means of bringing a representation into joint work: participants
use it to relate what has been said to the organization of activity and
to discover discrepancies that cannot be established merely by comparing
verbal formulations. For language AI, this distinction determines the
object of development. Correlating words with other words or with a
model's latent features does not yet disclose their connection with a
situation of activity. The machine must be able to reconstruct and
construct a representation of that situation, cut out from systemic
representations of activity, and to form a message on its basis. What is
then checked is the preservation in the utterance of the processes,
structures, and connections described. It is precisely this organization
that makes the foundations of a message available for joint examination
and transformation.

Boroday considers meaning in use to result from its construction in
working memory: ``Semantics receives its dynamic embodiment in working
memory, where meaning is constructed. Such construction takes place on a
verbal basis''.\footnote{Boroday, Язык и познание {[}Language and
  Cognition{]}, \emph{supra} note 80, at 583.} He relates the selection
of activated content to context: ``Which particular information enters
working memory upon activation depends on a number of factors:
linguistic, physical, and social context, background knowledge, the
characteristics of communication in a given community, and
others''.\footnote{Boroday, Язык и познание {[}Language and
  Cognition{]}, \emph{supra} note 80, at 583.} At the same time,
``long-term memory stores and continually updates an array of diverse
and poorly structured information, while working memory realizes the
dynamic interaction of linguistic form, sensorimotor systems, and
numerous cognitive operations''.\footnote{Boroday, Язык и познание
  {[}Language and Cognition{]}, \emph{supra} note 80, at 584.}

Boroday separates successful communication from completeness of
understanding: ``The answer is simple: successful communication does not
require exact translation and complete understanding''.\footnote{Boroday,
  Язык и познание {[}Language and Cognition{]}, \emph{supra} note 80, at
  593.} Discussing differences between the speaker's and listener's
representations, he specifies: ``Does this mean that the listener will
not fully understand the speaker? If understanding means complete
reconstruction of the mental model, then yes. However, it does not
follow that communication has not occurred''.\footnote{Boroday, Язык и
  познание {[}Language and Cognition{]}, \emph{supra} note 80, at 593.}
He describes the foundation of such interaction as follows: ``If we
analyze typical communicative situations even within a single language,
we will see that they unfold and are maintained on the basis of core
information, whereas the periphery remains underspecified and therefore
becomes the prerogative of individual imagination''.\footnote{Boroday,
  Язык и познание {[}Language and Cognition{]}, \emph{supra} note 80, at
  593.}

In the author's view, these propositions describe immature communication
and the work of an immature, non-technologized intellect, for which the
foundations of understanding remain implicit and the coordination of
content depends on participants' situational associations.

Reproducing ordinary human speech is insufficient if the task is to
ensure technologically organized strong thinking. A machine may form
expected phrases from context while a person supplies their content from
their own representations. Communication will occur, but the work of
reconstructing the situation and maintaining its organization will
remain with the person. Situational verbal success in such cases does
not attest to the machine's ability to work with the representations on
whose basis a message is formed and understood.

Strong AI must be able to construct and transform systemic
representations of activity while preserving within those
representations the distinctions between processes and structures and
the connections between them. The similarity of its responses to
immature, non-technologized human communication therefore does not
justify the machine's limitations; rather, it shows which way of working
the machine reproduces. Context dependence and underspecification
observed in ordinary human speech cannot, in themselves, serve as a norm
for such AI. Its adequacy must be judged by its ability to perform
organized work with the representations necessary for forming a message
and understanding it in joint activity.

\subsection*{3.2. Testing the Architectural Proposal}
\phantomsection
\label{yale-heading-008}
\addcontentsline{toc}{subsection}{3.2. Testing the Architectural Proposal}

The next research step is to implement a limited subject-domain example
of joint work with AI trained to reconstruct the five levels of
communicative content, and to compare response generation using it with
ordinary generation. Testing must separately assess the establishment of
the accepted foundation and the consequences of changing it,
reconstruction of the schema-model and the ability to work with it,
preservation of the situation's content in the message, and the
recipient's understanding. What matters is the preservation of
participants, negation, conditions, uncertainty, cost, and temporal
distinctions, as well as the consistency of the components of the full
sense.

The proposed experimental test uses existing tasks with underspecified
conditions. Ambig-SWE and Dialogue SWE-Bench make it possible to
investigate how a language agent identifies missing information, obtains
it through dialogue with a user, and uses it to solve a programming
task.\footnote{S. Vijayvargiya et al., \emph{Ambig-SWE: Interactive
  Agents to Overcome Underspecificity in Software Engineering},
  arXiv:2502.13069 (2025) (coauthors: X. Zhou, A. Yerukola, M. Sap, G.
  Neubig), \url{https://arxiv.org/abs/2502.13069}.}\footnote{B. King \&
  J. Flanigan, \emph{Dialogue SWE-Bench: A Benchmark for Dialogue-Driven
  Coding Agents}, arXiv:2606.13995 (2026),
  \url{https://arxiv.org/abs/2606.13995}.} Both approaches use SWE-bench
Verified material,\footnote{N. Chowdhury et al., \emph{Introducing
  SWE-bench Verified}, OpenAI (2024),
  \url{https://openai.com/index/introducing-swe-bench-verified/}.} which
is based on real SWE-bench tasks.\footnote{C. E. Jimenez et al.,
  \emph{SWE-bench: Can Language Models Resolve Real-World GitHub
  Issues?}, ICLR, arXiv:2310.06770 (2024) (coauthors: J. Yang, A.
  Wettig, S. Yao, K. Pei, O. Press, K. Narasimhan),
  \url{https://arxiv.org/abs/2310.06770}.} QuestBench makes it possible
to test separately whether an agent can identify a missing condition and
select a question to establish it.\footnote{B. Z. Li et al.,
  \emph{QuestBench: Can LLMs ask the right question to acquire
  information in reasoning tasks?}, NeurIPS, arXiv:2503.22674 (2025)
  (coauthors: B. Kim, Z. Wang), \url{https://arxiv.org/abs/2503.22674}.}
This material can be used to compare ordinary response generation with
the five-level organization of content by evaluating reconstruction of
the situation and the message's sense, the quality of clarifications,
and task solution. Authorial corrections, the use of corrected
distinctions in new tasks, and the revision of representations when
processes, structures, and their connections change are tracked
separately.

\begin{samepage}
\section*{Conclusion}
\phantomsection
\label{yale-heading-009}
\addcontentsline{toc}{section}{Conclusion}

The five-level communication architecture for strong language-based ASI
makes the foundations of an utterance explicit and available for
examination and adjustment, instead of using so-called ``context'' and
``weights'' as the foundation of a response. It connects acquired
schemas and systemic connections with cutting out a situation, message
formation, and understanding, and makes it possible to trace the
consequences of changing a foundation. The proposal also includes
reverse work: reconstructing, from a message, the situation and the
systemic foundations from which it was cut out, while checking the
connections preserved, lost, and changed. The next step for LLMs is the
reconstruction of large systemic models of activity, as distinct from
successful text and code generation.

\par\end{samepage}

\clearpage\section*{DECLARATIONS}\phantomsection\label{yale-declarations}\addcontentsline{toc}{section}{DECLARATIONS}

\par\vspace{16pt}\noindent{\endmatterfont\fontsize{11.5pt}{14pt}\selectfont Disclosure of AI Assistance\par}\vspace{8pt}\noindent\phantomsection
\label{yale-heading-010}

AI-assisted tools were used for language editing, structural review, and
citation audit support during manuscript preparation. The author remains
solely responsible for the conceptual analysis, source selection,
citation verification, and final text.

\par\vspace{16pt}\noindent{\endmatterfont\fontsize{11.5pt}{14pt}\selectfont Declaration of Competing Interests\par}\vspace{8pt}\noindent
\phantomsection
\label{yale-heading-011}

The author declares that they have no competing financial interests or
personal relationships that could have appeared to influence the work
reported in this paper.

\par\vspace{16pt}\noindent{\endmatterfont\fontsize{11.5pt}{14pt}\selectfont Funding\par}\vspace{8pt}\noindent\phantomsection
\label{yale-heading-012}

This research did not receive any specific grant from funding agencies
in the public, commercial, or not-for-profit sectors.

\par\vspace{16pt}\noindent{\endmatterfont\fontsize{11.5pt}{14pt}\selectfont Author's Statement on Defensive Publication\par}\vspace{8pt}\noindent\phantomsection
\label{yale-heading-013}

The author publishes the five-level architecture for language AI
described in this article as a defensive disclosure intended to
establish a public prior-art record and prevent subsequent patenting of
the disclosed combination.

\par\vspace{16pt}\noindent{\endmatterfont\fontsize{11.5pt}{14pt}\selectfont Acknowledgments\par}\vspace{8pt}\noindent
\phantomsection
\label{yale-heading-014}

I am grateful to my mother, Valentina Osovskaya, a linguist and a
student of Alexey Fyodorovich Losev (1893--1988)---the philosopher,
philologist, and scholar of antiquity who wrote on the philosophy of
language, the name, and the symbol---for the love of words, language
scholarship, linguistics, and the history of languages that she kindled
in me from childhood.

\end{document}